%% file: neurips_2025.tex
\documentclass{article}

\PassOptionsToPackage{numbers, compress}{natbib}
\usepackage{lmodern}
\usepackage{graphicx}
\usepackage[final]{neurips_2025}

\usepackage[table]{xcolor}
\usepackage[utf8]{inputenc} 
\usepackage[T1]{fontenc}    
\usepackage{hyperref}       
\usepackage{url}            
\usepackage{booktabs}       
\usepackage{amsfonts}       
\usepackage{nicefrac}       
\usepackage{microtype}      
\usepackage{xcolor}         

\usepackage{caption}
\usepackage{subcaption}
\usepackage{multirow}
\usepackage{dblfloatfix}

\title{Rope to Nope and Back Again: A New Hybrid Attention Strategy}

\author{Bowen Yang$^{1}$ \quad Bharat Venkitesh$^{1}$ \quad Dwarak Talupuru$^{1}$ \\ \textbf{Hangyu Lin$^{1}$ \quad David Cairuz$^{1}$  \quad Phil Blunsom$^{1}$ \quad Acyr Locatelli$^{1}$}\\\\
$^{1}$ Cohere \\\\
{\tt\small \{bowen, dwarak, john, phil\}@cohere.com\quad \{bharat, davidcairuz, acyr\}@cohere.ai}
}

\begin{document}

\maketitle

\input{abstract}
\input{introduction}
\input{observation}
\input{method}
\input{experiments}
\input{results}

\input{conclusion}

\bibliographystyle{abbrvnat}
\bibliography{ref}




\newpage
\appendix
\input{appendix}

\end{document}

%% file: abstract.tex
\begin{abstract}
Long context large language models (LLMs) have achieved remarkable advancements, driven by techniques like Rotary Position Embedding (RoPE) \cite{su2023roformerenhancedtransformerrotary} and its extensions \cite{chen2023extendingcontextwindowlarge, liu2024scalinglawsropebasedextrapolation, peng2023yarnefficientcontextwindow}. By adjusting RoPE parameters and incorporating training data with extended contexts, we can train performant models with considerably longer input sequences. However, existing RoPE-based methods exhibit performance limitations when applied to extended context lengths. This paper presents a comprehensive analysis of various attention mechanisms, including RoPE, No Positional Embedding (NoPE), and Query-Key Normalization (QK-Norm), identifying their strengths and shortcomings in long context modeling. Our investigation identifies distinctive attention patterns in these methods and highlights their impact on long context performance, providing valuable insights for architectural design. Building on these findings, we propose a novel architecture featuring a hybrid attention mechanism that integrates global and local attention spans. This design not only surpasses conventional RoPE-based transformer models with full attention in both long and short context tasks but also delivers substantial efficiency gains during training and inference.
\end{abstract}

%% file: introduction.tex
\section{Introduction}

Developing language models capable of handling long context lengths poses several challenges. First, as the context length increases, an effective modeling of extended input sequences becomes increasingly critical. This often requires advancements in positional encoding \cite{su2023roformerenhancedtransformerrotary}, extrapolation techniques \cite{ding2024longropeextendingllmcontext}, or architectural innovations \cite{tworkowski2023focusedtransformercontrastivetraining, huang2024advancingtransformerarchitecturelongcontext}. Second, training long context large language models with billions of parameters demands significant computational resources. Overcoming this challenge requires scalable algorithms, high-quality datasets, and robust infrastructure. Lastly, deploying these models in real-world applications demands low latency and low memory usage, which requires meticulous optimization of both the model architecture and the serving infrastructure. 

On the modeling front, two components of the transformer architecture are particularly crucial for long context capabilities: the attention mechanism and positional embeddings. Recent research has proposed various methods to enhance these components. For instance, Landmark Attention \cite{mohtashami2023landmarkattentionrandomaccessinfinite} trains attention modules to select relevant blocks using a representative token, referred to as a ``landmark token", for efficient retrieval within extended text corpora. Similarly, Focused Transformer \cite{tworkowski2023focusedtransformercontrastivetraining} adopts a contrastive training approach to prioritize attending the most relevant portions of the input sequence, allowing the model to focus on smaller, contextually significant subsets of tokens. Although these approaches improve long context modeling, stabilizing training remains a key challenge for extending transformer capabilities to longer sequences. Query-Key Normalization (QK-Norm) \cite{henry-etal-2020-query, rybakov2024methodsimprovingllmtraining} has been introduced to address the stability issue, which normalizes the query-key vectors along the head dimension before computing attention. Although QK-Norm mitigates numerical instability during training and is widely used \cite{chameleonteam2024chameleonmixedmodalearlyfusionfoundation, dehghani2023scalingvisiontransformers22, li2024hunyuanditpowerfulmultiresolutiondiffusion}, it may impair long context capabilities. 

In addition to the chosen attention mechanism, positional embeddings play a crucial role in long context modeling. Various approaches have been proposed to improve their effectiveness. Popular methods include Absolute Position Embedding (APE) \cite{vaswani2017attention}, Relative Position Embedding \cite{raffel2023exploringlimitstransferlearning}, ALiBi \cite{press2022trainshorttestlong}, and Rotary Position Embedding (RoPE) \cite{su2023roformerenhancedtransformerrotary}. Among these, RoPE has gained significant adoption in large language models (LLMs) \cite{dubey2024llama3herdmodels, yang2024qwen2technicalreport, aya_paper} due to its simplicity and effectiveness. In particular, it has the ability to extrapolate context lengths by adjusting RoPE $\theta$ values during training \cite{liu2024worldmodelmillionlengthvideo, ai2024yiopenfoundationmodels, aya_expanse_paper}. Other techniques, such as relative bias \cite{press2022trainshorttestlong, raffel2023exploringlimitstransferlearning, chi2022kerplekernelizedrelativepositional} and contextualized position embeddings \cite{golovneva2024contextualpositionencodinglearning}, introduce distance-based bias terms or condition the position information on input semantics. These methods often affect attention distribution by incorporating auxiliary information, such a positional indices or explicit recency bias. However, whether certain information or biases are beneficial to long context modeling or overall performance remains less explored. Additionally, the concept of No Positional Embedding (NoPE) has been explored by \cite{kazemnejad2023impactpositionalencodinglength}, suggesting that removing explicit positional embeddings and relying solely on implicit positional information derived from the causal mask can enhance long context performance. 

Despite advances in long context modeling, training and serving such models remain challenging due to the quadratic complexity of standard attention. Techniques like Sliding Window Attention (SWA) \cite{jiang2023mistral7b, gemmateam2024gemma2improvingopen} mitigate this by restricting each token’s attention to a local window, reducing compute while maintaining local coherence. Sparse attention methods \cite{child2019generatinglongsequencessparse, beltagy2020longformerlongdocumenttransformer, tay2020sparsesinkhornattention} extend this by introducing structured sparsity, including random \cite{zaheer2021bigbirdtransformerslonger} and dilated patterns \cite{ding2023longnetscalingtransformers1000000000}. More recent approaches further compress attention, such as activation beacon \cite{zhang2024longcontextcompressionactivation}, which sequentially summarizes local keys and values, and attention sink \cite{xiao2024efficientstreaminglanguagemodels}, which stabilizes long-sequence training by preserving early tokens with a sliding window. On the serving side, KV cache trimming methods \cite{zhang2023h2oheavyhitteroracleefficient, li2024snapkvllmknowslooking, cai2024pyramidkvdynamickvcache} selectively discard cached states based on heuristics to reduce memory usage and boost inference efficiency. However, these gains often come with trade-offs in model quality, emphasizing the need for careful design.

In this paper, we begin analyzing attention patterns of different attention mechanisms, RoPE, NoPE, and QK-Norm and its impacts on long context performance trained up to 750 billion tokens. Building on these insights, we propose a novel hybrid attention architecture and extensively pretrain up to 5 trillion tokens, followed by supervised fine-tuning on a diverse set of datasets tailored for long context. We show that this architecture surpasses existing state-of-the-art extrapolation-based RoPE models \cite{liu2024scalinglawsropebasedextrapolation} by a large margin, striking a balance between efficiency and performance.

%% file: observation.tex
\section{Observation}

In this section, we assess three models with different attention components on needles-in-a-haystack \cite{kamradt2023needle} (NIAH) and analyze the attention patterns to understand how these variants affect performance. Analysis in this section guides our architectural design choices throughout this work.

\subsection{Experimental Setup}
\label{sec:interleaved_training}

All model variants share a common configuration consisting of 8 billion parameters (including the token embedding parameters), with detailed architectural specifications provided in Table \ref{table:model_arch}. For this set of experiments, the model is trained in two stages: a pretraining stage followed by a supervised fine-tuning (SFT) stage. Previous research shows that the SFT stage is necessary for long context evaluations, as it can reduce variance in long context tasks and enables the emergence of long context capabilities that may not manifest in models trained solely through pretraining \cite{gao2024trainlongcontextlanguagemodels}.


We pretrain the model with a batch size of 4 million tokens. We use AdamW with a peak learning rate of $7e^{-3}$, a linear warmup of 2000 steps and a cosine learning rate schedule decaying to $3.5e^{-4}$ over 179,000 steps for a total of 750 billion tokens. For the SFT stage, we adopt an interleaved training strategy: we combine short- and long context data in a 3:1 ratio, with context lengths of 8192 and 65536 tokens, respectively. We use a batch size of 0.5 million tokens.

\begin{table}[!ht]
    \centering
    \fontsize{11pt}{15pt} \selectfont
    \makebox[\textwidth][c]{ 
    \begin{tabular}{cccccccc}
        \hline
        Parameters & Emb. Dim & FFN Dim & Non-linearity & Layers & Heads & KV Heads & Vocab Size \\
        \hline
        Values & 4096 & 28672 & swiglu & 32 & 32 & 8 & 256000 \\
        \hline
    \end{tabular}
    }
    \vspace{0.5em} 
    \caption{Model architecture details}
    \label{table:model_arch}
\end{table}


The 3 model variants we test are:

\begin{enumerate}

\item \textbf{RoPE Model:}
For this variant, we employ Rotary Position Embedding (RoPE) to encode positional information. During the pretraining stage, the RoPE parameter $\theta$ is set to 10,000. In the subsequent supervised fine-tuning (SFT) stage, $\theta$ is increased to 2 million to account for the increased context length. This variant serves as the baseline model configuration, maintaining an architecture similar to that of most existing models \cite{dubey2024llama3herdmodels, jiang2023mistral7b, cohere_for_ai_2024}.

\item \textbf{QK-Norm Model:}
Layer normalization \cite{ba2016layernormalization} is applied to both the query and the key vectors before performing the angular rotation used in RoPE. All other hyperparameters, including the $\theta$ values and training methodology, remain identical to those of the RoPE variant. 

\item \textbf{NoPE Model:} 
Previous research \cite{wang2024lengthgeneralizationcausaltransformers, haviv-etal-2022-transformer} has demonstrated that transformer variants trained without positional embeddings (NoPE) can perform effectively on long context tasks. However, these models often exhibit inferior performance in terms of perplexity and downstream task evaluations within the trained sequence length \cite{haviv-etal-2022-transformer}. In our study, the NoPE variant does not have QK-Norm. This variant is trained using the same methodology as the other two variants.

\end{enumerate}

\subsection{Evaluation and Attention Analysis}
\label{sec:NIAH_section_2}

\subsubsection{Evaluations}

We evaluate the variants on a set of core evaluation benchmarks, including MMLU \cite{hendrycks2021measuringmassivemultitasklanguage}, HellaSwag \cite{zellers2019hellaswagmachinereallyfinish}, CommonsenseQA \cite{talmor2019commonsenseqaquestionansweringchallenge}, ARC \cite{clark2018thinksolvedquestionanswering} for core model capabilities and NIAH benchmark \cite{kamradt2023needle} for long context capability. NIAH evaluates a model's ability to retrieve information accurately from a specific sentence (the ``needle'') embedded within a lengthy document (the ``haystack''). The needle is randomly placed at varying depths within the sequence to examine performance across different context lengths. To improve robustness, we modify the original NIAH benchmark, where each context-depth combination is tested 16 times with different random seeds, creating diverse context compositions for comprehensive evaluation. The results of all standard benchmark evaluations and results with 65536 context length needles are presented in Table \ref{tab:observation_first_eval}. Although prior research has emphasized the limitations of NIAH \cite{xu2024stresstestinglongcontextlanguagemodels} for evaluating deeper and more general context understanding, our focus is solely on testing basic long context capabilities and gaining insights on model architecture design, for which this benchmark is sufficient. 

Table \ref{tab:observation_first_eval} shows that the RoPE and QK-Norm variants exhibit comparable performance on standard benchmarks, while the NoPE variant lags behind. This finding is consistent with previous studies \cite{kazemnejad2023impactpositionalencodinglength, wang2024lengthgeneralizationcausaltransformers}. For long context evaluations, QK-Norm performs the worst among the three variants, despite its decent performance in other capabilities. This observation is consistent with the results from the comparisons between Command R and Command R+, where Command R, despite being a significantly smaller model, outperforms Command R+ overall on longer context benchmarks \cite{hsieh2024rulerwhatsrealcontext}. Although the NoPE variant has slightly lower needles score compared to the RoPE variant, it is decent given that its base capabilities is relatively low.

\subsubsection{Attention Pattern Analysis}
\label{sec:attention_pattern_analysis}

To better understand the impacts of different architectures, we also analyze the attention patterns within each model. This approach is inspired by previous studies \cite{ye2024differentialtransformer, leviathan2024selectiveattentionimprovestransformer} where attention scores assigned to different parts of the context are closely examined.

We still utilize the NIAH task by first dividing the context into four segments: 

\begin{itemize}

\item \textbf{Begin: }The first 10 tokens. This part of the context is also often referred to as the ``attention sink'' \cite{xiao2024efficientstreaminglanguagemodels}, where a disproportionately large amount of attention is typically allocated in a trained transformer model.

\item \textbf{Needle: }The tokens of the inserted needle sentence. Ideally, a well-trained model should assign a relatively high amount of attention to this part of the context.

\item \textbf{End:} The query and completion tokens, which can represent recency bias.

\item \textbf{Context:} The remainder of the context, typically consisting of noise or irrelevant content.

\end{itemize}

\begin{figure*}[t]
    \centering
    \begin{subfigure}[b]{\textwidth}
        \makebox[\textwidth][c]{%
            \includegraphics[width=1.1\textwidth]{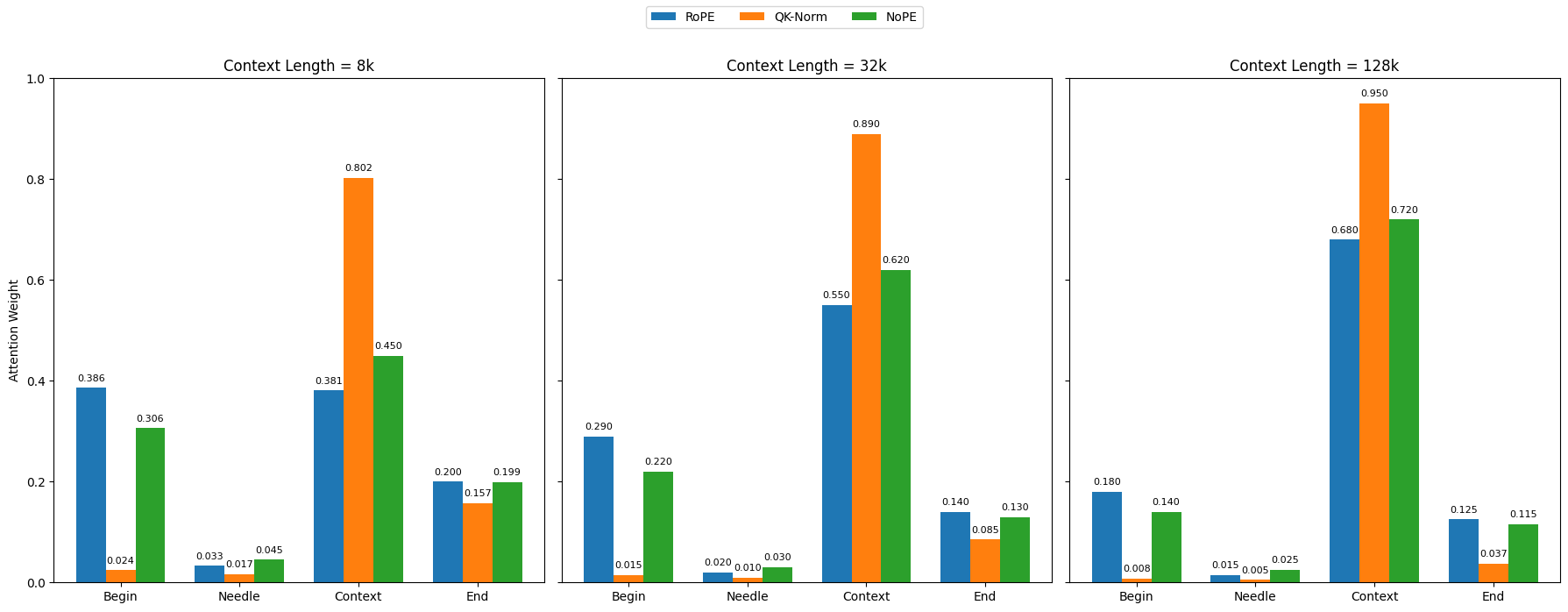}
        }
    \end{subfigure}
    \caption{Attention Distribution across context lengths of each variant}
    \label{fig:observation_attn_dist}
\end{figure*}

We position the needles at approximately 50\% depth within the context to increase the complexity of the task, as most models suffer from the lost-in-the-middle problem, as highlighted in previous works \cite{liu-etal-2024-lost, baker2024lostmiddleinbetweenenhancing}. For each model, we first calculate the attention scores between the query tokens of ``End'' segment and the key tokens of all four segments across all heads and layers. The attention scores are summed within each segment and then aggregated across all heads and layers to obtain the average attention weight for each segment. These scores are further averaged across multiple samples at sequence lengths of 8,000 tokens, 32,000 tokens, and 128,000 tokens. We refer to this metric as ``attention mass'' in the following sections. The results of each variant are visualized in Figure \ref{fig:observation_attn_dist}.

We begin by comparing attention distributions across different sequence lengths from Figure \ref{fig:observation_attn_dist}. We observe a consistent decrease in attention mass allocated to the ``Needle'' segment and a corresponding increase in attention mass on the ``Context'' segment as sequence length increases. This trend indicates that retrieving relevant information becomes increasingly difficult as the context grows longer. When comparing across model variants, the NoPE variant consistently allocates the highest attention mass to the Needle'' segment, followed by the RoPE variant, while the QK-Norm variant assigns the least attention to this segment. Furthermore, the QK-Norm variant exhibits markedly lower attention mass on the ``Begin'' segment and substantially higher attention mass on the ``Context'' segment relative to other variants. This indicates that models trained with the QK-Norm component exhibit a weak attention sink and are more susceptible to interference from noisy content. These patterns are consistent with QK-Norm’s poor performance on the NIAH task. We argue that QK-Norm has this effect because the normalization operation mitigates magnitude information from the dot product of Query and Key vectors which tends to result in attention logits being closer in terms of magnitude and flatter in terms of distribution. A more detailed analysis of why QK-Norm is detrimental to long-context modeling is provided in Appendix~\ref{appendix:B}.

\begin{table*}[!t]
    \centering
    \fontsize{11pt}{15pt} \selectfont
    \makebox[\textwidth][c]{ 
    \begin{tabular}{cccccccc}
        \hline
        Model & Val Loss & MMLU & HellaSwag & CommonsenseQA & ARC-E & ARC-C & Needles 65k \\
        \hline
        RoPE & 1.52 & 48.55 & 73.74 & 68.30 & 81.05 & 39.13 & 9.82 \\
        QK-Norm & 1.53 & 48.21 & 73.68 & 68.23 & 80.54 & 38.98 & \cellcolor{red!25}7.93 \\ 
        NoPE & \cellcolor{red!25}1.58 & \cellcolor{red!25}47.61 & \cellcolor{red!25}72.16 & \cellcolor{red!25}66.42 & \cellcolor{red!25}76.94 & \cellcolor{red!25}37.12 & 9.03 \\
        \hline
    \end{tabular}
    }
    \caption{Comparison of model performance across a range of benchmarks. All evaluations are based on the outputs of the SFT models. Red cells indicate lower performance.}
    \label{tab:observation_first_eval}
\end{table*}

\subsubsection{Hybrid Model}
\label{sec:division_of_labour}

Building on the findings above, we are inspired to combine RoPE and NoPE layer-wise to leverage the strengths of both variants. NoPE provides an effective attention mechanism for retrieving information based on vector similarity, while RoPE explicitly models positional information and introduces a recency bias. By combining them, we aim to enhance overall performance. We propose a new variant that alternates between NoPE and Rotary Position Embedding (RoPE) layers. Specifically, the two position-embedding strategies are interleaved, with NoPE applied in one layer and RoPE in the next. To ensure consistency and enable meaningful comparisons, the RoPE parameter $\theta$ is initially set to 10,000 during pre-training. The pre-training procedure follows the setup described in Section \ref{sec:interleaved_training} in terms of data and training protocol. Subsequently, we perform multiple fine-tuning runs with varying $\theta$ values—10,000, 100,000, 2 million, and 4 million—to evaluate performance across different configurations, using the same training steps and data as in Section \ref{sec:interleaved_training}. We refer to these models as RNoPE-10k, RNoPE-100k, RNoPE-2M, and RNoPE-4M with the specific RoPE $\theta$ value used for each during supervised fine-tuning (SFT).

\begin{figure*}[t]
    \centering
    \begin{subfigure}[b]{\textwidth}
        \makebox[\textwidth][c]{%
            \includegraphics[width=1.0\textwidth]{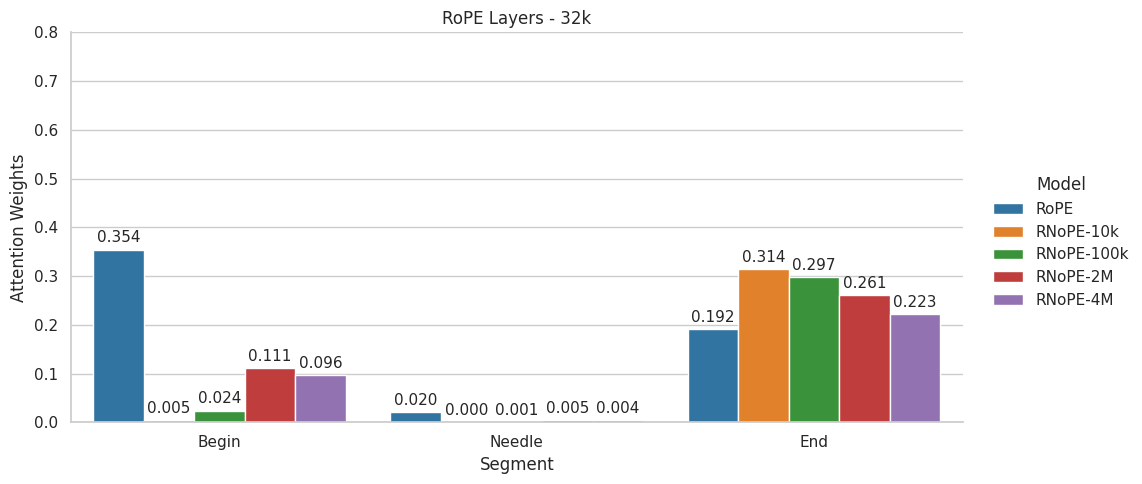}
        }
    \end{subfigure}
    \begin{subfigure}[b]{\textwidth}
        \makebox[\textwidth][c]{%
            \includegraphics[width=1.0\textwidth]{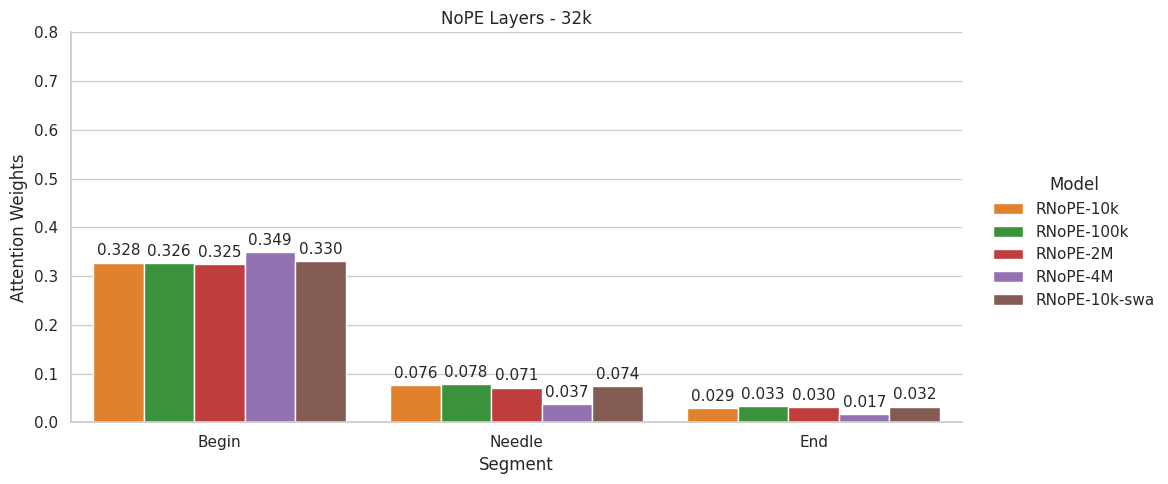}
        }
    \end{subfigure}
    \caption{Attention Distribution at 32k context length for rope baseline and hybrid variants.}
    \label{fig:observation_hybrid_model}
\end{figure*}


\begin{table}[!ht]
    \centering
    \fontsize{10pt}{15pt} \selectfont
    \makebox[\textwidth][c]{ 
    \begin{tabular}{ccccccc}
        \hline
        Model & RoPE & RNoPE-10k & RNoPE-100k & RNoPE-2M & RNoPE-4M & RNoPE-10k-swa \\
        \hline
        Needles-128k Score & 7.40 & 8.04 & 7.46 & 7.02 & 6.20 & 9.56 \\
        \hline
    \end{tabular}
    }
    \vspace{0.5em} 
    \caption{Needles score at 128k for all model variants}
    \label{table:hybrid_model_needles_score}
\end{table}

For the RoPE baseline model and all RNoPE variants, we report the needles evaluation score of all SFT models at a sequence length of 128,000 in Table \ref{table:hybrid_model_needles_score}. We also display the attention mass of all variants in Figure \ref{fig:observation_hybrid_model}. The attention mass is aggregated separately for all RoPE and NoPE layers. For simplicity, we present results based on samples with 32,000 sequence lengths, with the complete table for all sequence lengths provided in Appendix \ref{appendix:A}.

The results in Table \ref{table:hybrid_model_needles_score} reveal that increasing the RoPE parameter $\theta$ during fine-tuning of the RNoPE variants—where NoPE and RoPE layers are interleaved—leads to a decline in the model's long-context capability. In contrast, previous studies on pure RoPE architectures \cite{baichuanrope, liu2024worldmodelmillionlengthvideo} have shown that using a larger RoPE $\theta$ during pre-training or fine-tuning enhances long-context performance and expands the effective attention receptive field. To investigate this discrepancy, we compare the attention mass across different model variants from Figure \ref{fig:observation_hybrid_model}.

First, when comparing the NoPE and RoPE layers across all RNoPE variants, we observe a clear divergence in their behavior. The NoPE layers exhibit strong retrieval capabilities, characterized by a pronounced spike in attention mass on the ``Needle'' segment and a phenomenon of attention sink \cite{xiao2024efficientstreaminglanguagemodels} on the ``Begin'' segment. Additionally, these layers show a significantly weaker recency bias indicated by the low attention mass on the ``End'' segment. In contrast, the RoPE layers within the RNoPE variants demonstrate extremely weak retrieval performance, evidenced by near-zero attention mass on the “Needle” segment and only modest attention on the “Begin” segment—indicating an absence of an attention sink. However, these RoPE layers exhibit a much stronger recency bias compared to the pure RoPE baseline. In summary, interleaving RoPE and NoPE layers induces a spontaneous “division of labor” phenomenon, where RoPE layers focus on local information aggregation and NoPE layers specialize in long-range retrieval. Remarkably, this functional specialization emerged naturally during training, without any explicit training constraints, data augmentation strategies, or specific loss function designs.

Next, we examine the RNoPE variants fine-tuned with different $\theta$ values. As $\theta$ increases, we observe a consistent decline in the recency bias of the RoPE layers. Specifically, the attention mass on the “End” segment drops from 0.314 in RNoPE-10k to 0.223 in RNoPE-4M, indicating a flatter attention distribution that reaches deeper into the context. This aligns with prior findings and theoretical analyses suggesting that larger $\theta$ values expand the effective receptive field of the attention mechanism \cite{baichuanrope}. However, our empirical results indicate that this expanded receptive field in the RoPE layers introduces noise into the overall architecture, which disrupts the NoPE layers’ ability to compute similarities and perform retrieval effectively. This degradation is reflected in both attention mass and task performance: the needle attention mass in the NoPE layers drops from 0.0765 to 0.0369, and the needle evaluation score decreases from 8.036 to 6.203 as $\theta$ increases from 10,000 to 4 million. These findings further underscore that the model spontaneously develops a “division of labor” mechanism during training, with distinct roles emerging between RoPE and NoPE layers.

From these observations, we draw the following insights:
\begin{enumerate}
\item
\textbf{Division of Labour:} Combining NoPE and RoPE layers yields complementary strengths, with each type naturally assuming specialized roles after training. NoPE layers are adapted to information retrieval, while RoPE layers become effective at modeling local context due to their inherent recency bias.
\item
\textbf{Potential Efficiency Gains:} Restricting the effective attention span of the RoPE layers in RNoPE models can reduce noise in the attention distribution and reinforce the functional specialization of each layer. Additionally, it lowers the computational cost (FLOPs) during training—particularly for longer context lengths—thereby improving training efficiency while maintaining or even enhancing performance.
\end{enumerate}

Guided by these insights, we fine-tune a new variant, RNoPE-10k-swa, where “swa” denotes sliding window attention. This modification imposes a hard limit on the attention span of RoPE layers, operationalizing the second insight above. Specifically, the sliding window size for RoPE layers is set to 8,192, while the NoPE layers retain full attention to support long-context retrieval. All other training configurations remain identical to the RNoPE-10k variant, including the use of $\theta = 10{,}000$. Evaluation results, presented in Table \ref{table:hybrid_model_needles_score}, show a marked improvement. The RNoPE-10k-swa variant achieves a needles-128k score of 9.562, surpassing both the baseline and the original RNoPE-10k model. Moreover, its NoPE layers exhibit a well-structured attention pattern with high attention mass on both the “Begin” and “End” segments from \ref{fig:observation_hybrid_model}, reflecting strong long-context retrieval capabilities.

%% file: method.tex
\section{Model Architecture}
\label{sec:model_architecture}

\begin{figure*}[t]
    \centering
    \begin{subfigure}[b]{\textwidth}
        \makebox[\textwidth][c]{%
            \includegraphics[width=0.5\textwidth]{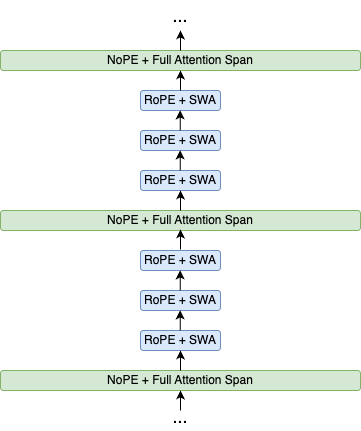}
        }
    \end{subfigure}
    \caption{RNope-SWA Model Architecture. SWA stands for sliding window attention.}
    \label{fig:model_arch}
\end{figure*}

Based on the analysis above, we make the following design choices on top of the Command R+ architecture \cite{cohere_for_ai_2024}. First, we remove the QK-Norm component due to its poorly shaped attention patterns, which adversely impacted long context performance. Second, NoPE layers with a full attention span are introduced to enhance the model's retrieval capabilities. Third, a sliding window size of 4,096 is applied to RoPE embeddings, leveraging RoPE's inherent recency bias to improve performance on short-to-medium context ranges. In particular, the sliding-window approach has been employed in several prior works \cite{gemmateam2024gemma2improvingopen, jiang2023mistral7b, noamcharacterpost}. Regarding the number of layers, we perform an ablation study on the interleaving ratio of full attention and sliding window layers, testing the configurations of 1:1, 1:3, and 1:7. The results show that a 1:3 ratio strikes an optimal balance between computational efficiency and performance. We position the full attention layers at the end of each interleaving group, preceded by three sliding window layers. All other hyperparameters for the model architecture remain consistent with those outlined in Table \ref{table:model_arch}. A visualization is shown in Figure \ref{fig:model_arch}.




For additional context, the resulting design shares similarities with other well-explored long-sequence architectures, such as Mega \cite{ma2023megamovingaverageequipped} and state-space models (SSMs) \cite{gu2022efficientlymodelinglongsequences, gu2024mambalineartimesequencemodeling}. For example, \cite{ma2023megamovingaverageequipped} and \cite{ma2024megalodonefficientllmpretraining} introduced a multi-dimensional damped version of an exponential moving average component in conjunction with the gated attention unit (GAU) architecture \cite{hua2022transformerqualitylineartime}, aiming to balance local and long-term dependencies, a common challenge in time-series modeling. Similarly, previous studies have proposed a diagonal variant of the S4 architecture \cite{gu2022efficientlymodelinglongsequences}, incorporating an exponentially decaying measure to enhance the model's ability to capture long-range dependencies \cite{tay2020longrangearenabenchmark}.

Earlier transformer variants, such as GPT-J \cite{gpt-j} and GPT-NeoX \cite{black-etal-2022-gpt}, explored hybridizing RoPE and NoPE by applying rotational embeddings to a subset of the head dimensions. The work of p-RoPE \cite{barbero2025roundroundgomakes} further advanced this line of research by analyzing RoPE in the frequency domain, revealing that low-frequency components capture semantic relationships while high-frequency components encode positional information. Similar to prior partial-RoPE approaches, p-RoPE removes selected low-frequency components along the head dimensions. These findings align with our observation that NoPE primarily facilitates retrieval (semantic focus), whereas RoPE exhibits a stronger recency bias (positional focus). However, prior works did not identify the cross-layer patterns underlying the “division of labor” phenomenon described in Section~\ref{sec:division_of_labour}, which enables our approach to achieve both improved performance and practical engineering benefits.


%% file: experiments.tex
\section{Experiments}

In this section, we detail the experiments conducted on the model architectures, covering the stages of pretraining, cooldown, supervised fine-tuning, and evaluations. Alongside long context evaluations, we provide a comprehensive assessment of short-context benchmarks, a dimension often disregarded in other long context studies. We train two models: one with the RoPE architecture as a baseline and another employing the architecture introduced in Section \ref{sec:model_architecture}.

\textbf{Pretraining and Cooldown.} The models are pretrained for 5 trillion tokens of diverse data with batch size of 8 million tokens using FP8 precision format \cite{micikevicius2022fp8formatsdeeplearning}. We use a cosine learning rate schedule of 5e-3 peak learning rate and 5\% end learning rate with 8,000 linear warmup steps. From the pre-trained model, we linearly anneal the learning rate from 2.5e-4 to 1e-6 for 50,000 steps in BF16 precision. The context length was initially maintained at 8k for the first 35,000 steps, then extended to 32k and 128k for 10,000 steps and 5,000 steps, respectively. For the baseline model, the RoPE $\theta$ values were increased to 1,000,000 for 32k and 8,000,000 for 128k contexts during the length extrapolation phase, while remaining constant for the RNope-SWA variant. Both models utilized the same interleaved training strategy outlined in Section \ref{sec:interleaved_training}.

\textbf{Supervised Finetuning.} We supervise fine-tune on top of the pretrained models. As the primary focus of this study is to evaluate the impact of architectural design on downstream tasks, preference training is deferred to future work. To preserve the long context capabilities of the model, the fine-tuning process utilizes interleaved datasets containing 8k and 128k prompt-response pairs. The long context SFT data at 128k includes Retrieval-Augmented Generation (RAG) \cite{lewis2021retrievalaugmentedgenerationknowledgeintensivenlp} datasets, multilingual translation datasets with extended passages, long code instruction datasets, and long context retrieval datasets. Training was performed for two epochs across the entire dataset.

%% file: results.tex
\section{Experimental Results}

Our evaluation contains a comprehensive analysis over standard benchmarks below 8k context length such as MMLU \cite{hendrycks2021measuringmassivemultitasklanguage}, HellaSwag \cite{zellers2019hellaswagmachinereallyfinish}, ARC \cite{clark2018thinksolvedquestionanswering}, SAT \cite{zhong2023agievalhumancentricbenchmarkevaluating}, GSM8k \cite{cobbe2021trainingverifierssolvemath}, Winograde \cite{sakaguchi2019winograndeadversarialwinogradschema} and MBPP \cite{austin2021programsynthesislargelanguage}, as well as popular long context benchmarks with needles-in-a-haystack \cite{kamradt2023needle} and the retrieval and QA portion of Ruler \cite{hsieh2024rulerwhatsrealcontext}. We test the context lengths up to 256k so we can examine the impact of these choices in the extrapolation capability of the model. We denote the baseline model trained with RoPE  scaling as ``Baseline" and the architecture with interleaved attention span and position embeddings as ``RNope-SWA".

\subsection{Standard Benchmarks}


In this set of evaluations, we evaluate baseline and RNope-SWA on a standard LLM benchmark covering various language, math and code capabilities. The results are shown in Table \ref{tab:short_evals}. We observe that the model is better or on-par on most of the metrics compared to the baseline and has some gains over the baseline numbers on certain benchmarks (+2.0\% on MMLU and +1.8\% on GSM8k). This set of results also indicates that although RNope-SWA explicitly removed position embeddings from 25\% of all its layers, positional information is retained by the interleaving attention span and captured by RoPE from previous layers. RNope-SWA does not have the performance loss due to the removal of explicit position embeddings, as previous works have shown \cite{kazemnejad2023impactpositionalencodinglength, wang2024lengthgeneralizationcausaltransformers}.

\begin{table*}[!t]
    \centering
    \fontsize{10pt}{15pt} \selectfont
    \makebox[\textwidth][c]{
    \begin{tabular}{cccccccccc}
        \hline
        Model &  MMLU & HellaSwag & ARC-E & ARC-C & SATEn & SATMath & GSM8K & Winogrande & MBPP \\
        \hline
        Baseline & 57.5 & 75.8 & \bf{84.6} & 48.5 & 70.0 & \bf{30.9} & 40.9 & 68.5 & 39.1 \\
        RNope-SWA & \bf{59.5} & \bf{76.2} & 82.5 & \bf{48.8} & \bf{71.9} & 30.5 & \bf{42.7} & \bf{69.5} & \bf{39.3} \\ 
        \hline
    \end{tabular}
    }
    \caption{Comparison of models on a variety of benchmarks. All the evaluations are based on the performance of the SFT-models.}
    \label{tab:short_evals}
\end{table*}

\subsection{Long Context Benchmarks}

To evaluate the long context performance of these models, we use NIAH and the retrieval and QA components of Ruler \cite{hsieh2024rulerwhatsrealcontext}. To better understand how architectural choices affect long context performance, we also evaluate with context lengths extending beyond training sequence length. This allows us to assess how well these models can interpolate as well as extrapolate to unseen context lengths and how specific architectural decisions influence these capabilities.


\subsubsection{NIAH Evaluations}

Following the settings of section \ref{sec:NIAH_section_2}, we run NIAH test 256k context lengths. The scores are reported in Table \ref{table:final_needle_eval} . The figure indicates that although both models are able to get close to perfect scores below the context length seen during training, RNope-SWA has better extrapolation capabilities and achieves almost no loss up to 256k context length while the baseline fails to extrapolate well -- despite using a high RoPE $\theta$ value of 8 million. We attach the graphical visualization with depth and length dimension expanded in Appendix \ref{appendix:C}.

\begin{table*}[!t]
    \centering
    \fontsize{10pt}{15pt} \selectfont
    \makebox[\textwidth][c]{
    \begin{tabular}{ccc}
        \hline
        Model &  Needles-128k & Needles-256k \\
        \hline
        Baseline & 9.99 & 8.25 \\
        RNope-SWA & 9.99 & 9.97 \\ 
        \hline
    \end{tabular}
    }
    \caption{Needles Score of Baseline and RNoPE-SWA up to 256k sequence length.}
    \label{table:final_needle_eval}
\end{table*}


\subsubsection{Ruler Evaluations}

\begin{table}[!ht]
	\centering
    \fontsize{10pt}{13pt} \selectfont
    \begin{tabular}{ccccccc}
        \hline
        Model &  8k & 16k & 32k & 64k & 128k & 256k \\
        \hline
        Baseline & \bf{96.6} & 94.4 & \bf{95.1} & 89.1 & 83.0 & 57.1 \\
        RNope-SWA & 96.1 & \bf{96.1} & 94.9 & \bf{92.0} & \bf{90.0} & \bf{74.8} \\ 
        \hline
    \end{tabular}
    \caption{Ruler Retrieval Evaluation}
    \label{tab:ruler_retrieval}
\end{table}

\begin{table}[!ht]
	\centering
    \fontsize{10pt}{13pt} \selectfont
    \begin{tabular}{ccccccc}
        \hline
        Model &  8k & 16k & 32k & 64k & 128k & 256k \\
        \hline
        Baseline & 53.5 & 50.0 & 52.5 & 45.5 & 36.0 & 30.0 \\
        RNope-SWA & \bf{55.5} & \bf{52.5} & \bf{55.5} & \bf{49.0} & \bf{46.0} & \bf{42.5} \\ 
        \hline
    \end{tabular}
    \caption{Ruler QA Evaluation}
    \label{tab:ruler_qa}
\end{table}

First introduced in \cite{hsieh2024rulerwhatsrealcontext}, the Ruler benchmark aims to provide a set of more difficult tasks than NIAH. It covers a wider range of retrieval under a Multi-Query/Key/Value settings, more realistic tasks with a long-context Question-Answering format and more. Although our modification of NIAH introduced more context variants and proves to be more difficult than the vanilla version, it still cannot test the limits of the model. Therefore, we evaluate our models on the retrieval and QA portion of the Ruler so we can better separate their performance.

From the results, we can observe that the baseline model with RoPE $\theta$ scaling approach suffers from a sharp drop in the longer context range, especially 64k and longer. Comparing the difference between the scores obtained at 8k and 256k, the baseline model dropped from 96.6 to 57.1 (about 41\%) on retrieval and from 53.5 to 30.0 (about 44\%) on QA, whereas the RNope-SWA model dropped from 96.1 to 74.8 (about 22.1\%) on retrieval and from 55.5 to 42.5 (about 23.4\%) on QA respectively. From the original Ruler Paper \cite{hsieh2024rulerwhatsrealcontext}, models that adopt similar RoPE scaling approaches have shown a similar degradation over longer context lengths \cite{dubey2024llama3herdmodels, yang2024qwen2technicalreport, abdin2024phi3technicalreporthighly} as the baseline.

\subsection{Impacts on Training and Inference}

We also report the differences in training and inference speed, as well as memory requirements, of RNope-SWA compared to the baseline model. Let $S$ denote the sliding window size and $L$ represent the full training context length. During training, 75\% of all layers now operate with a computational complexity of $\mathcal{O}(\textit{S} \textit{L})$, rather than the quadratic complexity of $\mathcal{O}(L^2)$. This results in the model being approximately 50\% faster than the baseline at a 64k context length and nearly 2x faster at 128k in terms of training throughput, using flash attention \cite{dao2022flashattentionfastmemoryefficientexact, dao2023flashattention2fasterattentionbetter, shah2024flashattention3fastaccurateattention} and a sequence-parallel scheme similar to \cite{jacobs2023deepspeedulyssesoptimizationsenabling, yang2024contextparallelismscalablemilliontoken}. With alternative implementations, such as Ring Attention \cite{liu2023blockwiseparalleltransformerlarge, liu2023ringattentionblockwisetransformers} or its variants \cite{brandon2023stripedattentionfasterring}, sliding window adoption can reduce key-value block communication overhead if carefully sharded along the sequence dimension, potentially improving speed. In theory, KV cache memory and time complexity of RNoPE-SWA can be reduced by up to 75\%. Empirically, we observed a 44\% latency reduction at 132k input and 96 output tokens, and nearly 70\% at 990k input and 8 output tokens—approaching the theoretical limit as sequence length grows. Increasing the ratio of sliding window to full attention layers can further improve speed and memory efficiency.


%% file: conclusion.tex
\section{Discussions and Future Work}

In this paper, we introduced RNope-SWA, an architecture that interleaves NoPE and RoPE position embeddings with varying attention spans (RNope-SWA). RNope-SWA is able to strike a balance between effective attention modeling and computational efficiency, achieving a nearly 4x reduction in KV cache size and significantly boosting both training and inference speeds without compromising performance. The integration of NoPE layers with full attention spans enhances long context capabilities, eliminating the need for RoPE scaling. This simplification improves the stability of training and delivers excellent long context performance.

Our findings align with recent work, such as YoCo \cite{sun2024cacheoncedecoderdecoderarchitectures}, Jamba-1.5 \cite{jambateam2024jamba15hybridtransformermambamodels}, and MiniMax-01 \cite{minimax2025minimax01scalingfoundationmodels}, which demonstrate that hybrid attention mechanisms generally outperform full attention mechanisms in handling long contexts. However, the underlying reasons behind this seemingly counterintuitive observation remain largely unexplored. This opens an intriguing area of study, particularly as models push towards multi-million-token context lengths. Re-visiting and re-thinking the foundational components of transformer architectures, such as attention mechanisms, may become essential to accommodating these extreme requirements. Recent works \cite{ye2024differentialtransformer, leviathan2024selectiveattentionimprovestransformer, soh2024usereactiveattentionslice} have begun to explore this direction by focusing on reducing attention noise across large context windows, a promising approach to refine the performance of attention modules.


%% file: appendix.tex
\section{Attention Distribution of All Lengths}
\label{appendix:A}

This table contains the attention distribution of RoPE and RNoPE variants from Section \ref{sec:attention_pattern_analysis} over 8k, 32k and 128k sequence lengths.

\begin{table*}[ht]
\centering
\fontsize{11pt}{15pt} \selectfont
\makebox[\textwidth][c]{
\begin{tabular}{cccccccccc}
\hline
\multirow{2}{*}{Context Length} & \multirow{2}{*}{Model} & \multicolumn{4}{c}{NoPE Layers} & \multicolumn{4}{c}{RoPE Layers} \\ 
                       &                               & Begin & Needle & Context & End  & Begin & Needle & Context & End  \\ \hline
                       
\multirow{6}{*}{8k} & RoPE & - & - & - & - & 0.3863 & 0.0328 & 0.3809 & 0.2000 \\
                     & RNoPE-10k & 0.3900 & 0.0952 & 0.4736 & 0.0412 & 0.1255 & 0.0102 & 0.5340 & 0.3302 \\
                     & RNoPE-100k & 0.3854 & 0.0932 & 0.4783 & 0.0430 & 0.2135 & 0.0136 & 0.4558 & 0.3171 \\ 
                     & RNoPE-2M & 0.3775 & 0.0902 & 0.4874 & 0.0449 & 0.2041 & 0.0126 & 0.4952 & 0.2881 \\
                     & RNoPE-4M & 0.4153 & 0.0546 & 0.5072 & 0.0229 & 0.1389 & 0.0136 & 0.6162 & 0.2313 \\
                     & RNoPE-10k-swa & 0.3830 & 0.1025 & 0.4702 & 0.0443 & 0.2040 & 0.0110 & 0.5938 & 0.1911 \\
                     \midrule
\multirow{6}{*}{32k} & RoPE & - & - & - & - & 0.3541 & 0.0201 & 0.4343 & 0.1915 \\
                     & RNoPE-10k & 0.3275 & 0.0765 & 0.5672 & 0.0287 & 0.0049 & 0.0004 & 0.6805 & 0.3142 \\
                     & RNoPE-100k & 0.3263 & 0.0778 & 0.5633 & 0.0327 & 0.0241 & 0.0005 & 0.6782 & 0.2972\\ 
                     & RNoPE-2M & 0.3250 & 0.0712 & 0.5735 & 0.0303 & 0.1111 & 0.0046 & 0.6233 & 0.2611 \\
                     & RNoPE-4M & 0.3486 & 0.0369 & 0.5981 & 0.0165 & 0.0960 & 0.0039 & 0.6774 & 0.2227 \\
                     & RNoPE-10k-swa & 0.3303 & 0.0742 & 0.5634 & 0.0321 & - & - & - & - \\
                     \midrule
\multirow{6}{*}{128k} & RoPE & - & - & - & - & 0.3463 & 0.0010 & 0.4751 & 0.1776 \\
                     & RNoPE-10k & 0.2991 & 0.0444 & 0.6430 & 0.0135 & 0.0000 & 0.0001 & 0.7230 & 0.2769 \\
                     & RNoPE-100k & 0.2454 & 0.0419 & 0.7016 & 0.0111 & 0.0001 & 0.0000 & 0.7749 & 0.2250 \\ 
                     & RNoPE-2M & 0.2600 & 0.0401 & 0.6836 & 0.0162 & 0.0417 & 0.0008 & 0.7516 & 0.2059 \\
                     & RNoPE-4M & 0.2949 & 0.0307 & 0.6635 & 0.0109 & 0.0663 & 0.0022 & 0.7115 & 0.2230 \\
                     & RNoPE-10k-swa & 0.2760 & 0.0467 & 0.6615 & 0.0159 & - & - & - & - \\
                     \bottomrule
                     
\end{tabular}
}
\caption{Needles Attention Pattern: RoPE and RNoPE variants}
\label{table:observation_first_attn}
\end{table*}

\section{Attention Distribution of RoPE and QK-Norm variants}
\label{appendix:B}

In this section, we further investigate the suboptimal performance of the QK-Norm variant. We present three plots comparing the attention distribution between the RoPE and QK-Norm variants across sequence lengths of 8k, 32k, and 128k on needle samples, following the setup outlined in Section~\ref{sec:attention_pattern_analysis}. Additionally, we provide the aggregated attention entropy for each variant to quantitatively support the arguments.

\begin{figure*}[t]
    \centering
    \begin{subfigure}[b]{0.49\textwidth}
        \includegraphics[width=\textwidth]{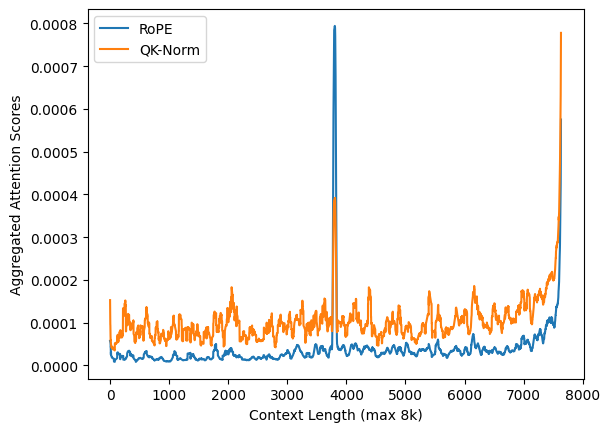}
        \caption{Context Length 8k}
        \label{fig:fig7}
    \end{subfigure}
    \hfill
    \begin{subfigure}[b]{0.49\textwidth}
        \includegraphics[width=\textwidth]{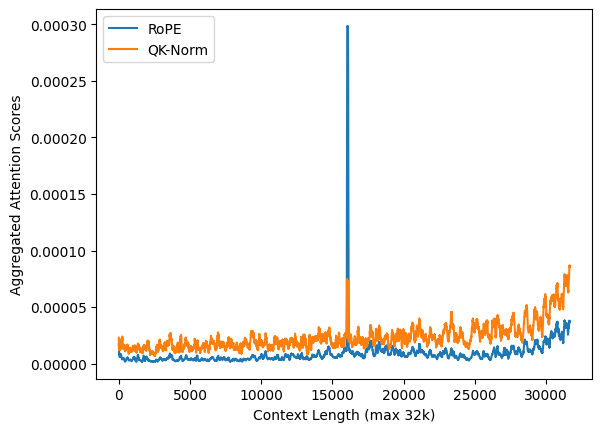}
        \caption{Context Length 32k}
        \label{fig:fig8}
    \end{subfigure}
    \hfill
    \begin{subfigure}[b]{0.49\textwidth}
        \includegraphics[width=\textwidth]{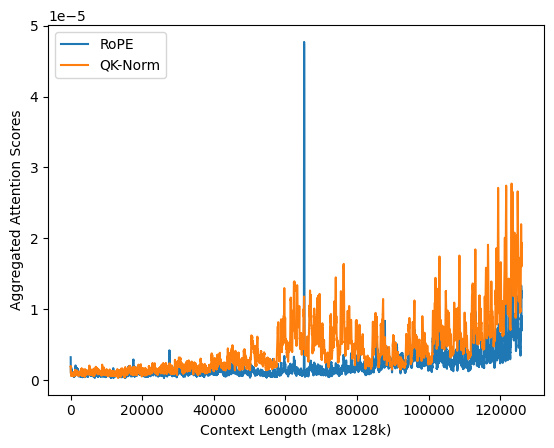}
        \caption{Context Length 128k}
        \label{fig:fig9}
    \end{subfigure}
    \hfill
    \caption{Attention Distribution Across Sequence lengths}
    \label{fig:attention_distribution_appendix}
\end{figure*}

To enhance the clarity of the distribution plots, we preprocess the attention distribution array by removing the first 10 tokens and the last 3\% of tokens from each sequence. This preprocessing step mitigates the disproportionate attention mass resulting from the attention sink effect and the recency bias observed in RoPE, thereby making the attention patterns more interpretable. We then compute a moving average with a window size of 100 tokens and average the results across all samples and layers to generate the final distributions.

\begin{center}
    \begin{table*}[ht]
    	\centering
        \fontsize{10pt}{15pt} \selectfont
    	\begin{tabular}{cccccccccc}
    		\hline
            Model &  8k & 32k & 128k \\
    		\hline
    		RoPE & \bf{6.02} & \bf{6.95} & \bf{7.62} \\
    		QK-Norm & 10.71 & 12.46 & 14.14\\ 
            \hline
        \end{tabular}
        \caption{Entropy values of aggregated attention distribution}
        \label{tab:entropy}
    \end{table*}
\end{center}


From Figure \ref{fig:attention_distribution_appendix}, we observe that the QK-Norm variant exhibits a lower spike on needle tokens but distributes more attention mass across context tokens. However, it also demonstrates a stronger recency bias compared to the RoPE variant. This characteristic results in a lower signal-to-noise ratio for the QK-Norm variant, which hampers its ability to effectively retrieve relevant information from long contexts. To further quantify this observation, we calculate the entropy values of the attention distributions for both variants, averaging across samples and layers at each sequence length. The results, listed in Table \ref{tab:entropy}, show that the QK-Norm variant has significantly higher entropy values than the RoPE variant. This aligns with its weaker performance in long context retrieval tasks, as higher entropy reflects a more dispersed and less focused attention distribution.

\section{Needles Score at 256k}
\label{appendix:C}

\begin{figure*}[t]
    \centering
    \begin{subfigure}[b]{\textwidth}
        \makebox[\textwidth][c]{%
            \includegraphics[width=1.2\textwidth]{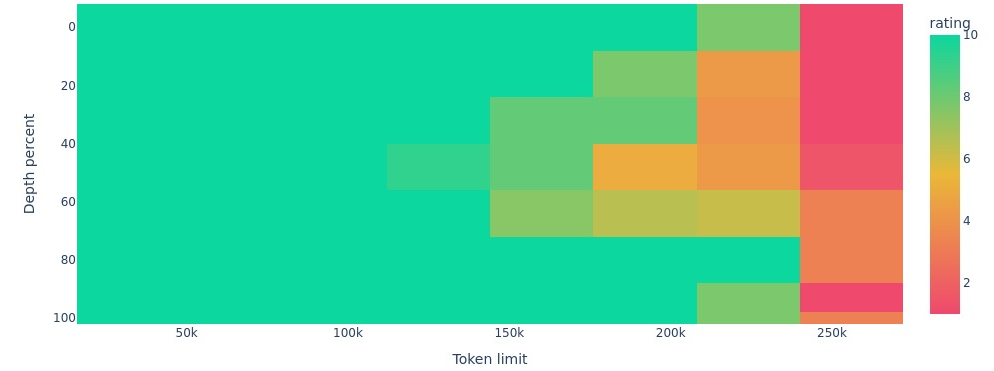}
        }
        \caption{Baseline}
    \end{subfigure}
    \begin{subfigure}[b]{\textwidth}
        \makebox[\textwidth][c]{
            \includegraphics[width=1.2\textwidth]{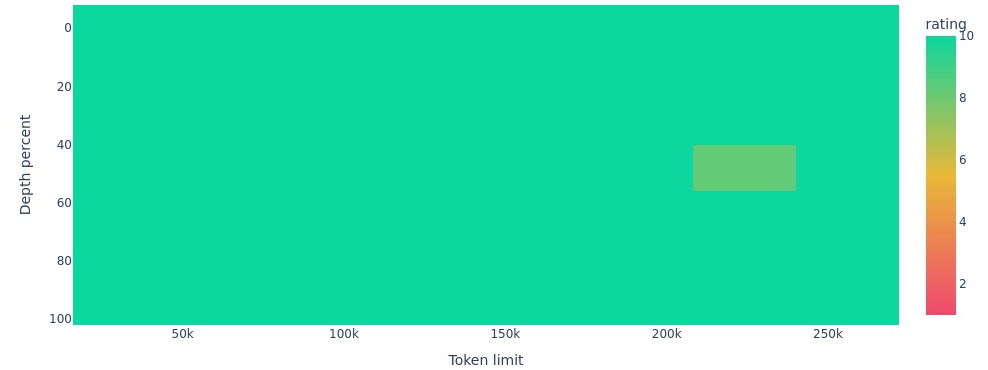}
        }
        \caption{RNoPE-SWA}
    \end{subfigure}
    \caption{Needle Evaluation of Baseline and RNoPE-SWA on 256k sequence length}
    \label{fig:final_needle_eval}
\end{figure*}